\documentclass[11pt,a4paper]{article}
\usepackage[hyperref]{acl2021}
\usepackage{times}
\usepackage{latexsym}

\usepackage{hyperref}
\usepackage{enumitem} 
\usepackage[lofdepth,lotdepth]{subfig}
\usepackage{xcolor}
\usepackage{comment} 
\usepackage{amsmath, amssymb, amsthm, enumerate, framed, graphicx}
\usepackage{url}
\usepackage{wrapfig,lipsum,booktabs}
\usepackage{multirow}
\usepackage{cleveref}
\usepackage{CJKutf8}
\usepackage{soul}

\usepackage{microtype}

\aclfinalcopy 

\definecolor{mypink}{cmyk}{0, 0.7808, 0.4429, 0.1412}
\definecolor{mypurple}{rgb}{0.57, 0.36, 0.51}
\definecolor{myblue}{rgb}{0.2, 0.2, 0.6}

\title{Detecting Hallucinated Content in \\Conditional Neural Sequence Generation}

\author{Chunting Zhou$^1$\thanks{\;\;Most work was done during an internship at FAIR.}, Graham Neubig$^1$, Jiatao Gu$^2$, Mona Diab$^2$, Paco Guzman$^2$, 
\\\textbf{Luke Zettlemoyer}$^2$, \textbf{Marjan Ghazvininejad}$^2$ \\
Language Technologies Institute, Carnegie Mellon University$^1$\\
Facebook AI Research${^2}$ \\
\texttt{\{chuntinz,gneubig\}@cs.cmu.edu},\\
\texttt{\{jgu,mdiab,fguzman,lsz,ghazvini\}@fb.com}
}

\date{}

\begin{document}
\maketitle
\begin{abstract}
Neural sequence models can generate highly fluent sentences, but 
recent studies have also shown that
they are also prone to hallucinate additional content not supported by the input.
These variety of fluent but wrong outputs are particularly problematic, as it will not be possible for users to tell they are being presented incorrect content.
To detect these errors, we propose a task to predict whether each token in the output sequence is hallucinated (not contained in the input) and collect new manually annotated evaluation sets for this task.
We also introduce a method for learning to detect hallucinations using pretrained language models fine tuned on synthetic data that includes automatically inserted hallucinations  
Experiments on machine translation (MT) and abstractive summarization demonstrate that our proposed approach consistently outperforms strong baselines on all benchmark datasets. 
We further demonstrate how to use the token-level hallucination labels to define a fine-grained loss over the target sequence in low-resource MT and achieve significant improvements over strong baseline methods.%
We also apply our method to word-level quality estimation for MT and show its effectiveness in both supervised and unsupervised settings
\footnote{Codes and data available at \url{https://github.com/violet-zct/fairseq-detect-hallucination}.}.
\end{abstract}

\section{Introduction}
\label{sec:intro}
Neural sequence models for tasks such as data-to-text generation~\citep{puduppully2019data}, machine translation (MT; \citet{vaswani2017attention,wu2016google}) and text summarization \citep{rothe2020leveraging} can often generate fluent text that is sometimes \emph{preferred} to human-written content \citep{laubli-etal-2018-machine,brown2020language}.
However, they also often generate texts that lack global logical consistency~\citep{gary2020gpt3}, are dull and repetitive~\citep{welleck2019neural}, or contain hallucinated content that is not entailed by the input~\citep{maynez_acl20,martindale2019identifying}.
In this paper, we focus on tackling the latter problem, aiming to automatically identify and quantify content in the output that is not faithful to the input text.

\begin{figure}[t]
 \vspace{-2mm}
    \centering
      \includegraphics[width=0.45\textwidth]{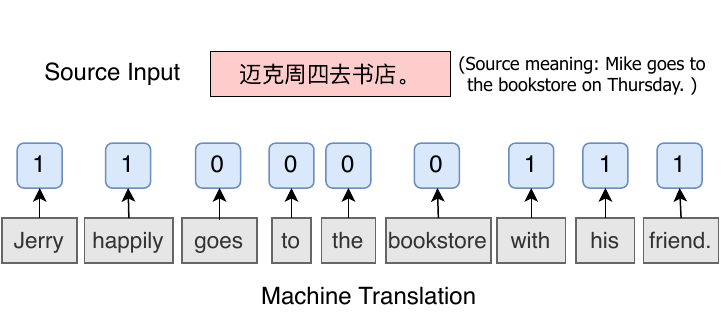}
       \vspace{-2mm}
    \caption{An example of token-level hallucination detection from MT. The grey box is an example of MT output and the labels above indicate if each word is faithful (0) to the input or hallucinated (1).}
    \label{fig:example}
    \vspace{-6mm}
\end{figure}

The risk of generating unfaithful content impedes the safe deployment of neural sequence generation models. 
The first step to building models that do not suffer from these failures is the assessment and identification of such hallucinated outputs.
Prior work has shown that standard metrics used for text evaluation, such as BLEU scores \citep{Papineni:2002,post-2018-call}, ROUGE \citep{lin:2004:ACLsummarization} and BERTScore \citep{zhang2019bertscore}, do not correlate well with the faithfulness of model outputs~\citep{maynez_acl20,wang-sennrich-2020-exposure,tian2019sticking}. They also require reference output text, limiting their applicability in a deployed system at run-time.
Very recent efforts have started to develop automatic metrics to measure the faithfulness of output sequences using external semantic models, e.g. the question-generation and question-answering systems~\citep{wang-etal-2020-asking,durmus-etal-2020-feqa} or textual entailment inference models~\citep{maynez_acl20}, to score faithfulness tailored for abstractive text summarization.
However, these scores do not directly identify hallucinated tokens
and only correlate weakly with human judgements.
   
We propose a new task for faithfulness assessment - hallucination detection at the token level, which aims to predict if each token in the machine output is hallucinated or faithful to the source input. 
This task does not use the reference output to assess faithfulness, which offers us the ability to also apply it at run-time.
Similar to the spirit of our proposed task, word-level quality estimation~\citep{specia2018findings,fonseca-etal-2019-findings} in the MT community predicts if tokens are correctly translated based on human post-editing. However, these methods generally do not distinguish errors in terms of fluency and adequacy~\citep{specia2011predicting}, with the exception of a subset of the WMT 2020 shared task on quality estimation~\citep{specia2020wmtqe}, where different types and levels of severity of word-level errors are defined.
Our proposed task specifically focuses on hallucination errors, and we define these errors in a simpler way with only binary labels, which we argue makes them simpler to use and more conducive to labeling at large scale.
The proposed hallucination detection method (described below) is also applicable to the word-level quality estimation task as demonstrated in \S\ref{sec:wqe}.

We measure hallucination for two conditional sequence generation tasks -- abstractive summarization and MT. For the former, we produce a benchmark dataset from recently released annotations ~\citep{maynez_acl20}. For MT, we carefully design human assessment guidelines and create high-quality annotations, which will be released to aid future research. To learn token-level hallucination prediction for general conditional sequence generations tasks, we propose a novel method that creates synthetic ``hallucinated" data and finetunes a pretrained language model~\citep{liu2019roberta,conneau-etal-2020-unsupervised} on it.
Without any human annotated supervised training data, we achieve an average F1 of around 0.6 across all the benchmark datasets, setting initial performance levels for this new task.


Predicting hallucination labels at the token level provides a tool for diagnosing and interpreting model outputs, which allows us to flag potential risks when the model is applied to previously unseen inputs.
Additionally, we show how to use these token-level hallucination labels 
in two case studies
to improve self-training~\citep{scudder1965probability} and learning from noisy mined bitext~\citep{koehn2019findings} in low-resource MT.
In both cases, there can be noise in the target text, either produced by the self-training teacher or errors in the mining process.
However, most outputs are only partially erroneous (see examples in Appendix \ref{app:st:examples}) and the rest of the output is still useful for training, as we show by introducing different token-level loss truncation schemes that use our proposed hallucination detection methods.
Our best methods outperform strong baselines by a large margin, and reduce the number of hallucinations.

\section{Token-level Hallucination Prediction}
\label{sec:task}



For source sequence $S$ and generated output sequence $G$, 
following ~\citet{maynez_acl20}
we define any span $g_i, \cdots, g_{i+j}~(j >= 0)$ in $G$ as being ``hallucinated'' if it is not supported by the source input $S$.%
\footnote{Content that is paraphrased or can otherwise be inferred by the source document is not considered hallucinated.}
More specifically, we consider two types of hallucination, which are not mutually exclusive:
\vspace{-1mm}
\paragraph{Extrinsic hallucinations:} a span $g_i, \cdots, g_{i+j}$ in $G$ consists of additional content without clear grounding in the input. In Fig.~\ref{fig:example}, the word \textit{``happily"} in the machine translation belongs to this case, as there is no word in the input sentence that clearly corresponds to ``happy''.
\vspace{-1mm}
\paragraph{Intrinsic hallucinations:} a span of word(s) in $G$ contains incorrect information due to synthesizing content using information present in $S$. 
In Fig.~\ref{fig:example}, \textit{``Jerry"} in the MT is a hallucinated word and should be replaced by \textit{``Mike"}.
Note that multi-word phrases can also be marked intrinsic hallucinations, such as ``this is a book'' being hallucinated from ``this is not a book'', where ``this is'' is a minimal span corresponding to the hallucination.

The above definitions are for illustrative purposes; we do not explicitly label whether a hallucination is intrinsic or extrinsic, only whether one exists at all.
Given these spans, we aim to identify all the span(s) satisfying the above conditions in machine generation $G$.\footnote{We do not annotate under-generations e.g. the source input is only partially translated or summarized.}

\vspace{-1mm}
\paragraph{Human Assessment of Hallucinations}
\label{sec:human:eval}
To facilitate the assessment of hallucinations in MT, we conduct human annotations on outputs of MT models in the patent and COVID-19 domain.
Three bilingual annotators were presented the source sentence, the reference sentence and the MT output, and they were asked to 
label each sentence with one of the three types of labels: incomprehensible, faithful, and contains hallucinations.
If the translation contains hallucinations, we asked the annotators to tag all the tokens that were not faithful to the source.
The final benchmark datasets were created by taking majority labels among three annotators.
We present more details regarding annotation guidelines and pipelines in Appendix \ref{appex:eval}.

We compute the Fleiss's Kappa~\citep{fleiss1971measuring} (FK) scores of our annotations for MT and the processed annotations from~\citep{maynez_acl20} on abstractive summarization (Tab.~\ref{tab:fleiss} in Appendix \ref{appex:eval}). We achieved moderate agreement (FK$\approx$0.56) on the token-level hallucination annotations and substantial agreement (FK$\approx$0.67) on the sentence-level annotations, while~\citet{maynez_acl20} achieved substantial or almost perfect agreement (FK$\approx$0.8) on the \textsc{XSum} dataset. For MT, we conjecture that it is relatively hard to achieve consistent agreement among annotators for several reasons.
First, although we have made detailed annotation guidelines following the definition of hallucination in \S~\ref{sec:task}, it could still be difficult for annotators to distinguish between ungrammatical translations and hallucinations.
Second, it was sometimes difficult for annotators to understand the specialized text in the patent domain.


\section{Token-level Hallucination Detection}
We propose a general-purpose method for token-level hallucination detection for conditional sequence generation tasks.
Given the source input $S$, we first formulate the task of token-level hallucination detection as a sequence labeling problem where a binary label is predicted at each position $G_t$ of the machine generation $G$. 
One straightforward way of learning this task is to train a model with supervised data in the form of $((S, G), L_G)$ where $L_G$ are the labels at every position of $G$ that indicate if each word is a hallucinated one or not.
However, because such labeled training data is not readily available, we propose an approach to automatically create synthetic training data.

\subsection{Synthetic Data Creation}
\label{sec:syn:data}
We use bi-text from the training data to create synthetic examples by automatically inserting new, hallucinated target-side tokens. 
More specifically, we take target sequence $T$ and create a hallucinated version of it denoted $T'$ with associated hallucination labels for each token in $T'$. Then we can train a supervised model on this synthetic labeled data set of $((S, T'), L_{T'})$. The key challenge is that $T'$ should be a fluent sentence that does not differ too much from $T$.

\vspace{-1mm}
\paragraph{Generation of hallucinated sentences} 

\begin{figure}[t]
    \centering
      \includegraphics[width=0.45\textwidth]{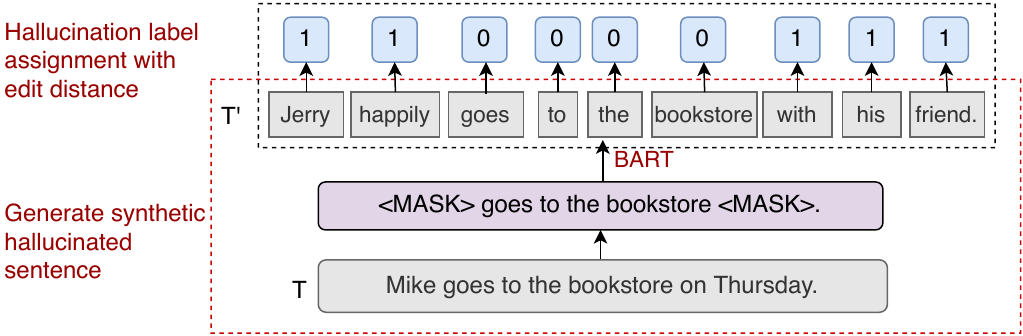}
    \caption{Generation of synthetic data with hallucination labels. A hallucinated version of $T$ is generated by feeding the noised sentence to the encoder-decoder model BART. Hallucination labels are assigned to each token by computing the edit distance between $T'$ and $T$. Labels of 1 refer to hallucinated words.}
    \label{fig:syn}
    \vspace{-5mm}
\end{figure}

To control this synthetic hallucination process, we build on a pre-trained denoising autoencoder, which maps a corrupted sentence back to the original text it was derived from, learning to reconstruct missing words that have been arbitrarily masked out. 
Specifically, we use the BART model \citep{lewis-etal-2020-bart}, without providing it any access to the source sentence, thereby encouraging it to insert new content as needed to ensure fluency.
As shown in Fig.~\ref{fig:syn}, we first apply a noising function that removes words from the original target sentence $T$\footnote{We also applied other noising functions, please see \S\ref{sec:exp:syn}} and then use a pretrained BART to generate $T'$ conditioned on the noised $T$ with beam search.

\begin{figure}[h]
\begin{center}
    \includegraphics[width=0.45\textwidth]{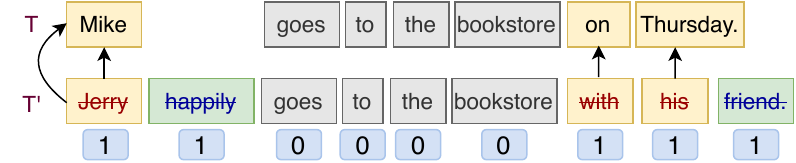}
    \caption{An example of label assignment.}
    \label{fig:edit}
\end{center}
\vspace{-6mm}
\end{figure}
\paragraph{Label assignments} After obtaining the hallucinated sentence $T'$ with BART, we need to assign appropriate labels to each token in $T'$ to mark which words are hallucinated. We compute the edit distance between $T'$ and $T$, and back-trace the deletion and substitution operations with dynamic programming. 
All the positions in $T'$ involving these two operations are labeled as hallucinations and everything else is considered faithful to $T$. 
Fig.~\ref{fig:edit} shows an example of label assignment with edit distance, where words in red are replaced and words in blue are deleted to convert $T'$ to $T$. 
Assigning labels with edit-distance can not always guarantee correct labels, but we find 
that this simple approach provides sufficiently high quality training data for effective hallucination detection in practice.

\subsection{Finetuning on Synthetic Data}
\label{sec:model:finetune}
\begin{figure*}[t]
    \centering
      \includegraphics[width=0.8\textwidth]{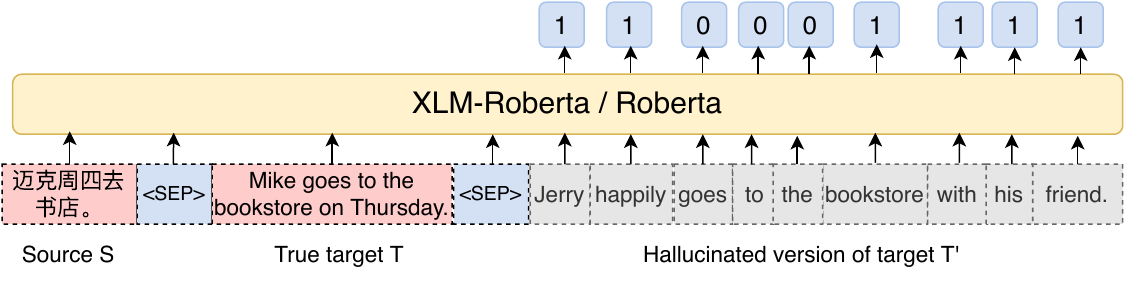}
      \vspace{-2mm}
    \caption{Finetuning XLM-Roberta (for cross-lingual generation task, e.g. MT) or Roberta (for monolingual generation task, e.g. text summarization) on the synthetic training data.}
    \label{fig:finetune}
    \vspace{-3mm}
\end{figure*}

\vspace{-1mm}
\paragraph{Hallucination prediction loss} We follow the common practice in natural language understanding (NLU) tasks and finetune a pretrained language model (LM) on our synthetic data.
We finetune a cross-lingual LM~\citep{conneau-etal-2020-unsupervised} for MT and a monolingual LM~\citep{liu2019roberta} for summarization.
In both cases, we concatenate the input, true target and hallucinated target denoted ($S$, $T$, $T'$) as a single input sequence to the model. Then we minimize the standard classification loss  $\mathcal{L}_{pred}$ over the pseudo hallucination labels $L_{T'}$ on top of the final hidden vectors of each token in $T'$ as shown in Fig.~\ref{fig:finetune}. 

Although using only the source text and hallucinated target ($S$, $T'$) as the input should be sufficient to learn to predict hallucinations, we can also easily measure the extent to which including the true target $T$ in the input could help the model. At test time, when evaluating the faithfulness of the machine outputs $G$, we do not use the true target $T$ and perhaps surprisingly find our model can generalize well without references, even when they were present during training.

To prevent the model from overly relying on the true target $T$ and learning spurious correlations (e.g. the edit distance), we explored two techniques: (1) \textit{dropout} -- randomly drop out tokens in $T$ to force the dependence on the source input; (2) \textit{paraphrase} -- recall that at synthetic data generation time, we generate $T'$ from BART conditioned on the noised $T$.
Instead, we can apply noise functions to the paraphrased sentence of $T$. 
We create paraphrased targets via knowledge distillation~\citep{kim2016sequence} where we use the output from pretrained Seq2Seq model conditioned on the source sentence in the bi-text corpus as the paraphrased target.
Let $D$ denote the paraphrased sentence of $T$ and $D'$ denote the generation from BART conditioned on the noised $D$.
Then we create pseudo labels of $D'$ denoted $L_{D'}$ by computing the edit-distance between the $D'$ and $D$ and use $((S, T, D'), L_{D'})$ as the training data for finetuning. 
Since the pseudo labels are created based on $D$, it can prevent the model from learning the edit-distance between $T$ and $D'$ easily.
We provide ablation studies in Appendix~\ref{sec:abl}. 

\vspace{-1mm}
\paragraph{Masked LM loss} We also add the masked language model loss (MLM) $\mathcal{L}_{mlm}$ following~\citep{devlin2019bert}. To learn this loss, we create a different batch from the above by concatenating only the source $S$ and target $T$ as the input, since the hallucinated target $T'$ could provide erroneous information for predicting masked words in $T$. We find that such multi-task learning objective helps learn better representations of the input and further improves performance on predicting hallucination labels. The final loss is $\mathcal{L} = \mathcal{L}_{pred} + \alpha \cdot \mathcal{L}_{mlm}$ where $\alpha$ is a hyperparameter.

\begin{table*}[t]
\centering
\small
\begin{tabular}{l|cc|cccc}
\toprule  
  \multirow{2}{*}{\textbf{Methods}}  & \multicolumn{2}{c|}{\textbf{MT}} & \multicolumn{4}{c}{\textbf{Summarization}} \\
 & \textbf{TranS2S} & \textbf{MBART} & \textbf{PtGen} & \textbf{TConvS2S} & \textbf{TranS2S} & \textbf{BERTS2S} \\
\midrule
Alignment & 29.47 & 9.93 & 38.92 & 37.94 & 34.47 & 35.81 \\
Overlap-based & 9.14 & 3.24 & 57.22 & 54.25 & 53.79 & 55.13  \\
Synonym-based & -- & -- & 59.54 & 63.73 & 58.66 & 53.07 \\
\midrule
Ours (w/o reference) & \textbf{65.75} & \textbf{41.92} & 63.66 & 65.94	& 61.70 & 55.45 \\
Ours (w/o reference + synonym) & -- & -- & \textbf{64.72} & \textbf{69.37}	& \textbf{63.88} & \textbf{56.49} \\
Ours (w/ reference) & 66.08 &	46.81 & 63.89 & 66.28 &	62.24 &	55.88\\
\bottomrule
\end{tabular}
\caption{F1 (x100) of hallucination labels on MT (see\cref{sec:data:mt}) and abstractive summarization (\textsc{XSum}). The first block are baseline methods and the second block are our results. Bold indicates best results not using references.}
\label{tab:token}
\vspace{-4mm}
\end{table*}

\begin{table*}
     \centering
     \small
\begin{tabular}{l|cc|cccc}
\toprule  
  \multirow{2}{*}{\textbf{Methods}}  & \multicolumn{2}{c|}{\textbf{MT}} & \multicolumn{4}{c}{\textbf{Summarization}} \\
 & \textbf{TranS2S} & \textbf{MBART} & \textbf{PtGen} & \textbf{TConvS2S} & \textbf{TranS2S} & \textbf{BERTS2S} \\
\midrule
        True hal. tokens (\%) & 18.12 & 11.10 & 46.09	& 52.89 &	46.74 &	37.51 \\
        Pred hal. tokens (\%)  & 18.56 & 7.99 & 57.22	& 57.68 &	55.78 &	48.84 \\
        \bottomrule
    \end{tabular}
    
    \caption{\label{tab:mbart:at} Annotated (True) and predicted (Pred) percentage of hallucinated tokens on benchmark test sets.}
    \vspace{-3mm}
    \label{tab:hal:tokens}
\end{table*}

\section{Evaluation Tasks and Data}
We examine hallucination in abstractive text summarization and machine translation (MT) tasks, using the models and datasets described below.

\subsection{Abstractive Text Summarization}
\label{sec:data:xsum}
\citet{maynez_acl20} studied hallucination problems in extreme summarization on the \textsc{XSum} dataset which comprises 226,711 British Broadcasting Corporation (BBC) articles paired with their single-sentence summaries. They randomly sampled 500 articles from the \textsc{XSum} test set and 
evaluated summaries from four abstractive summarization systems: \textbf{PtGen}~\citep{see2017get}, \textbf{TConvS2S}~\citep{xsum-emnlp}, \textbf{TranS2S}~\citep{vaswani2017attention} and \textbf{BERTS2S}~\citep{rothe2020leveraging}.
\citet{maynez_acl20} asked human annotators to label the spans in the machine generated summaries if they were unfaithful to the article.
We post-processed their human annotations by majority voting and created test datasets for each of the summarization systems.

\subsection{MT}
\label{sec:data:mt}
Previous work~\citep{wang-sennrich-2020-exposure,muller2019domain,koehn2017six} has shown that translation models are particularly prone to hallucination when tested out of domain. We similarly focus on this regime and additionally consider the low resource case where a modest amount of out of domain data is available at training time. 


\paragraph{Data} We use a multi-domain Chinese-English (Zh-En) translation dataset~\citep{wang2020aaai} which consists of four balanced domains: \textit{law, news, patent} and \textit{subtitles}. We create a new training data $\mathcal{D}_{train}$ with \textit{law} (1.46M sentences), \textit{news} (1.54M), \textit{subtitles} (1.77M) train data and randomly sample 870 parallel sentences from the \textit{patent} training data. 
We train two NMT models (described below) on this dataset and test on 150 examples from the patent test data. In addition, we also test the NMT models on the COVID-19 domain, sampling 100 examples from the dataset of \citet{anastasopoulos2020tico}.
We denote this 250-sentence dataset as $\mathcal{D}_{eval}$ and ask human annotators to evaluate the level of hallucinations thereof. 

\paragraph{Models} 
Our data is generated from two models on which we will measure hallucination (see Appendix~\ref{appex:exp} for more details):
(1) \textbf{TranS2S}~\citep{vaswani2017attention} is the standard Transformer Seq2Seq model with 6 encoder layers and 6 decoder layers.
(2) \textbf{MBART}~\citep{liu2020multilingual} is a Seq2Seq denoising auto-encoder pretrained on large-scale monolingual corpora in many languages.
We finetune the 12 layer model on $\mathcal{D}_{train}$. 

\section{Experiments}
\label{sec:exps}
\subsection{Experimental setup}
\label{sec:exp:syn}
\paragraph{Synthetic Data Generation} We use a pretrained 12 layer BART~\citep{lewis-etal-2020-bart} model in the fairseq toolkit~\citep{ott2019fairseq} for synthetic labeled data generation. 
We uniformly sample the percentage of tokens $p_m$ to mask from $[0, h_m]$ for each sentence.
We also uniformly sample the probability of replacing a token with a random token from $[0, h_r]$ denoted $p_r$. $p_m$ and $p_r$ are two important factors that affect the noise level when generating the synthetic data. For MT, we set $h_m$ and $h_r$ to 0.6 and 0.3 respectively. For abstractive summarization, we use 0.4 and 0.2. We use beam search for decoding from BART with beam size of 4 and length penalty of 3.
For MT, we first create paraphrased target sentences $D'$ through knowledge distillation~\citep{kim2016sequence} by using the outputs from the same trained TranS2S model on the source inputs.

\paragraph{Hallucination Predictor} For MT, we finetune XLM-R~\citep{conneau-etal-2020-unsupervised}
on the synthetic dataset with batch size of 128, and we annotated 50 examples (different from those in $\mathcal{D}_{eval}$) from the patent test data as the validation dataset.
For summarization, we finetune RoBERTa~\citep{liu2019roberta} with batch size of 96 and early stop training with 10K update steps. In addition, we dropout tokens from the reference $T$ in the input with a rate of $0.5$ and $0.3$ respectively for summarization and MT to learn $\mathcal{L}_{pred}$.
We set $\alpha$ to be 0.6 for MT and 0.5 for summarization based on the scales of $\mathcal{L}_{pred}$ and $\mathcal{L}_{mlm}$.
For both tasks, we set the mask probability used for $\mathcal{L}_{mlm}$ to be 0.5, and the initial learning rate to be $2e-5$ with polynomial decay.
We describe other hyperparameters, including training of MT models, in the Appendix~\ref{appex:exp} and \ref{app:exp:finetune}.

\subsection{Evaluation of hallucination prediction}
\label{sec:eval:token}

In Tab.~\ref{tab:token}, we present the F1 of token-level hallucination labels across six benchmark datasets for MT and abstractive summarization (full results of precision, recall and F1 are presented in Tabs.~\ref{tab:token:full:mt} and \ref{tab:token:full:xsum} in the appendix).
We compare with three baseline methods that we proposed for this new task: \textbf{(1)} The \textbf{alignment-based} method uses a word alignment model for hallucination assessment.  We employ SimAlign~\citep{sabet2020simalign}, an unsupervised aligner, that extracts word alignments from similarity matrices induced from pretrained word embeddings. SimAlign is essentially used for crosslingual tasks, and we adapt it to summarization by using embeddings from the pretrained BERT-large~\citep{devlin2019bert}. We predict a target token as being hallucinated if it is not aligned to the source tokens. \textbf{(2)} The \textbf{overlap-based} method is a heuristic one that predicts a target token as being hallucinated if does not appear in the source. Since it's not feasible to perform string matching between two languages for MT, we use a bilingual lexicon induction method~\citep{zhou19naacl} to first translate each English word into a Chinese word and then check its existence in the source text. \textbf{(3)} We go further by exploiting \textbf{synonyms} to assess hallucination in the summarization task where we use WordNet~\citep{miller1998wordnet} to find synonyms of nouns, verbs, adjectives and adverbs of the target summary and the source article; we predict a target as being hallucinated if its synonym can not be found in the set of the source synonyms.

From Tab.~\ref{tab:token}, we note:
\textbf{(1)} The proposed method achieves decent performance on this task and ranks the best among all baseline methods. However the task is still far from being solved
is worthy of study in the future.
\textbf{(2)} We can see that even though our model learns hallucination prediction with reference $T$ during training (Sec.~\ref{sec:model:finetune}), by applying token dropout to $T$, our model generalizes well without feeding the reference at test time. As a contrast, we report the results of predicting with reference at test time and observe that the model can achieve a significantly higher recall but worse precision (Tab.~\ref{tab:token:full:xsum} in appendix).
\textbf{(3)} The two non-neural baselines we proposed work surprisingly well on the summarization datasets, especially the synonym-based system. 
We guess this is because the information of the summaries should come from the source article and a majority of hallucinated words are nouns (\S\ref{sec:analysis}) which can be easily detected by string matching or synonym matching.
Our neural system performs better than these baseline methods but not significantly, and we hypothesize that this is because
the RoBERTa model we finetune on only allows a maximum input length of 512,  which results in an average cutoff of 158 subwords from the source article and hence loss of source information. 
By taking the union of the predictions from the synonym-based and our models, we can further obtain improvements on the summarization datasets.
We believe the advances in long sequence modeling~\citep{beltagy2020longformer,kitaev2020reformer} could help here, and are important to study in future work.
\textbf{(4)} At the same time, the baseline methods can not obtain reasonable performance for MT since crosslingual semantic matching is more challenging and our model shows significant improvements.

In Tab.~\ref{tab:hal:tokens}, we show the percentage of annotated and model predicted hallucinated tokens across the six benchmark sets. We can see that model predictions correlate well with human assessment and have a Pearson correlation coefficient of 0.986.

\begin{figure*}[t]
    \centering
      \includegraphics[width=0.85\textwidth]{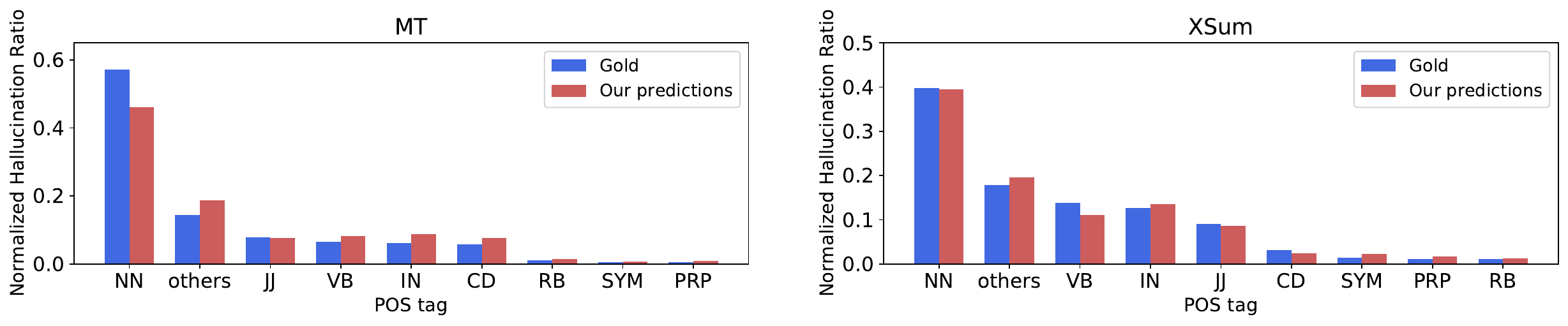}
      \vspace{-2mm}
    \caption{Relationship of POS tags and percentage of hallucinations for MT (left) and summarization (right).}
    \label{fig:macro:pos}
    \vspace{-3mm}
\end{figure*}

\subsection{Analysis}
\label{sec:analysis}

\paragraph{Analysis on Pretrained Models for Conditional Sequence Generation}
Recent work~\citep{maynez_acl20} has shown that pretrained models are better at generating faithful summaries as evaluated by humans.
In Tab.~\ref{tab:hal:tokens}, summaries generated from BERTS2S contain significantly fewer hallucinations than other model outputs.
We also confirmed this trend in MT that translations from MBART contain less hallucinated content than that from TranS2S.
\paragraph{Analysis on Hallucinated Words and their Part-of-Speech Tags}
In Fig.~\ref{fig:macro:pos}, we present the percentage of hallucinated tokens categorized by their part-of-speech tags predicted by a POS tagger~\citep{toutanova2003feature}. First, we see that for both MT and summarization datasets, nouns are the most hallucinated words. In abstractive summarization, verbs also account for a certain number of hallucinations.  Second, our model predicted hallucinated words match well with gold annotations on the distributions of POS tags. We also compare the percentage of hallucinations within each POS tag in Appendix \ref{app:micro:pos}. In addition, we provide more ablation studies in Appendix~\ref{sec:abl}.

\subsection{Evaluation on Word-level Quality Estimation}
\label{sec:wqe}
As noted in \S\ref{sec:intro}, our model is also readily applicable to word-level quality estimation (QE) for MT~\citep{fonseca-etal-2019-findings, specia2020wmtqe}, which aims to detect word-level errors in MT output.
In the WMT shared task of word-level QE, each token of the target sentence is labeled as OK/BAD based on the post-edited target sentences.
We evaluate our model on the WMT18 en-de word-level QE shared task~\citep{specia2018findings} in both the unsupervised and supervised setting.
There are 13,442 labeled parallel sentences where the tagged target sentences are from an NMT model.
In our supervised setting, we finetune the XLM-R model on these parallel sentences with the objective: $\mathcal{L}_{pred} + 0.5*\mathcal{L}_{mlm}$.
In the unsupervised setting, we first create the synthetic data (\S\ref{sec:syn:data}) using the post-edited target sentences from the labeled parallel set (13,442) and an additional 50K target sentences from the provided unlabeled parallel set.
Then we finetune XLM-R on the created synthetic labeled data.
For both settings, we set the weights of the cross-entropy loss for the bad-token labels to be 2.0 because the labels are imbalanced with fewer bad-token labels.
\begin{table}[h]
\small
\setlength{\tabcolsep}{3pt}
    \centering
    \begin{tabular}{l|ccc}
    \toprule
        Models & BAD-F1 & OK-F1 & F1-MULT \\
    \midrule
        OpenKiwi & - & - & 44.77 \\
        1$^{st}$ place in WMT18 & 48.00 & 91.00 & 44.00 \\
        3$^{rd}$ place in WMT18 & 36.00 & 85.00 & 30.00 \\
        \midrule
        Ours (unsupervised) & 37.09 & 92.54 & 34.32 \\
        Ours (supervised) & \textbf{50.78} & \textbf{91.91} & \textbf{46.68} \\
        \bottomrule
    \end{tabular}
    \caption{F1 scores (x100) on the test set of WMT18 word-level QE. OpenKiwi~\citep{kepler2019openkiwi} is the state-of-the-art result on this task. 1$^{st}$ and 3$^{rd}$ place are results from the shared task~\citep{specia2018findings}.}
    \label{tab:word:qe}
    \vspace{-4mm}
\end{table}

\paragraph{Results} We present results in Tab.~\ref{tab:word:qe}, where F1-Mult is the multiplication of F1-scores for the OK and BAD labels. 
Note that all the baseline models are in the supervised setup and the best baseline OpenKiwi~\citep{kepler2019openkiwi} is a strong ensembled system using predictions from multiple models. 
In contrast, our supervised model only leverages the parallel labeled data without using other resources.
Among all the supervised settings, our model outperforms the best system by 2 points in F1-Mult.
To make it clear how our unsupervised model performs, we also show the best performed systems in the shared task of WMT18. 
We observe that our unsupervised setting achieves descent performance and even outperforms the 3$^{rd}$-ranked system.
These results demonstrate that both the full version and the finetuning part of our method provide strong results for word-level QE. 

\section{Case Study I: Improving Self Training in Machine Translation}
\label{sec:exp:st}

Predicting hallucination labels at token-level 
not only allows us to flag potential risks in generation models, but also opens up the possibility of providing fine-grained signals which can be used to define new learning objectives. In this section and the following one, we demonstrate how to leverage the hallucination labels to reduce adverse effects of noisy training instances. Specifically, we show that the fine-grained hallucination signals allow for improved semi-supervised learning (\S\ref{sec:exp:st}) and training with noisy parallel data  (\S\ref{sec:exp:nf}).

\subsection{Rectified Self-Training for Neural MT}
Self training~\citep{scudder1965probability} is an important semi-supervised approach that utilizes unlabeled source data to improve system performance.
In a conditional sequence generation task, a teacher model is first trained with bitext $\mathcal{D}_l = \{\mathbf{s}_i, \mathbf{t}_i\}_{i=1}^{N}$ and used to make predictions on each sequence in a unlabeled dataset $\mathcal{D}_{u} = \{\mathbf{s}_j\}_{j=N+1}^{N+M}$ to create \textit{pseudo parallel data} $\mathcal{D}_p = \{\mathbf{s}_j, \mathbf{t'}_j\}_{j=N+1}^{N+M}$. The model is then trained on $\mathcal{D}_{l} \cup \mathcal{D}_p$.
\citet{he2019revisiting} finds that with self-training the student model can benefit from such pseudo-parallel data.
However, such results require a relatively high-quality teacher, and performance suffers in low-resource setting where no such teacher is available. 

We propose to use our token-level hallucination predictions to define a fine-grained loss during training in MT, by penalizing errors less on tokens that more likely to be hallucinated.
This is in contrast to previous data filtering methods for MT, which remove entire sentence pairs~\citep{junczys2018dual,kang-hashimoto-2020-improved}.

First, we predict the token-level hallucination labels on the target side of the pseudo parallel data $\mathcal{D}_p$.
Then we propose two simple methods of using these labels in self-training:
(1) We discard the losses of tokens that are predicted as hallucinations and compute the loss on the remaining tokens for each target sequence (\textbf{token loss truncation}). 
(2) Instead of adjusting losses, we mask the decoder hidden states of those hallucinated positions after the target-to-source cross attention in each decoder layer (\textbf{decoder HS masking}).\footnote{We also tried removing hallucinated target words before training. This underperformed, likely because it produces too many ungrammatical target sentences.}

\begin{table}[h]
\centering
\small
\setlength{\tabcolsep}{3pt}
\begin{tabular}{lccc}\\\toprule  
Methods & BLEU  &  BLERUT   & Hal (\%) \\\midrule
baseline & 16.14 & -0.166 & 13.69\\
ST & 19.31 & -0.059 & 10.00\\
\midrule
ST + paraphrase noise (ST-P) & 19.05 & -0.051 & 13.48 \\ 
ST + random noise (ST-R) & 19.97 & -0.041 & 12.55 \\
\midrule
ST + seq loss truncation & 19.91 & -0.048 & 8.26 \\
ST-R + seq loss truncation & 19.37 & -0.057 & 10.06\\ 
\midrule
ST + token loss truncation & 20.32 & 0.00244 & \textbf{6.37}\\
ST + decoder HS masking & 20.57 & -0.0001 & 6.38 \\
ST-R + token loss truncation & \textbf{21.02} & \textbf{0.043} & 7.34 \\
ST-R + decoder HS masking & 20.64 & 0.0308 & 8.70 \\
\bottomrule
\end{tabular}
\caption{BLEU($\uparrow$), BLEURT($\uparrow$) and hallucinated tokens (Hal,  $\downarrow$) on the CWMT2017 test set. We compare with noised self-training and sequence-level loss truncation in the second and third blocks respectively.}
\label{tab:data:aug}
\vspace{-2mm}
\end{table}

\subsection{Experimental Setup and Results}
\paragraph{Experimental Setup} To train a teacher model (baseline in Tab.~\ref{tab:data:aug}), we use the same training data described in \S\ref{sec:data:mt} using \textit{patent} (870) as the low-resource domain. We evaluate on the full patent test set (1,500) from CWMT2017~\citep{wang2020aaai}. For the unlabeled data, we use the withheld Chinese patent training data (2.9M). 

\paragraph{Baselines} We compare with the state-of-the-art self-training (ST) method of \citet{he2019revisiting}, which injects two types of noise into the input sentences: (1) paraphrase noise created by round-trip translations, and (2) random noise from dropping, masking and shuffling input tokens. We also compare with the recently proposed loss truncation method~\citep{kang-hashimoto-2020-improved} that adaptively removes entire examples with high log loss, which was shown to reduce hallucinations.

\paragraph{Results and Analysis} We present the tokenized BLEU score~\citep{Papineni:2002}, BLEURT score~\citep{sellam2020bleurt} and the percentage of hallucinated tokens predicted by our system in Tab.~\ref{tab:data:aug}. 
We can see that ST improves over the baseline by around 3 BLEU and our best result further improves ST by 1.7 BLEU.
Compared with strong baseline methods, our method not only achieves the best translation quality measured by BLEU and BLEURT but also the largest hallucination reduction. 
We also observe that: 
(1) Our method with ST alone can outperform other baseline methods,  when combined with perturbed ST (noise), and using fine-grained control over the target tokens can further improve the results. 
(2) ST with paraphrase noise (by round-trip translation) does not perform as well as the random noise, which further confirms that the noisy outputs from a teacher model may hurt the student model. 
(3) The sequence-level loss truncation approach can improve over the vanilla ST and reduce the level of hallucinations as measured by our system. However, the performance drops when combined with the noised ST.

\section{Case Study II: Improving Corpus Filtering for Low-Resource MT}
\label{sec:exp:nf}
High-quality parallel data is critical for training effective neural MT systems, but acquiring it can be expensive and time-consuming. Many systems instead use mined and filtered parallel data to train NMT models~\citep{junczys2018dual,zhang2020parallel,koehn2019findings}. 
Nonetheless, the selected parallel data can still be noisy, containing misaligned segments.
In this section, we demonstrate that token-level hallucination labels can allow us to make better use of noisy data to and improve the overall translation quality.
We apply the token loss truncation method proposed in \S\ref{sec:exp:st} to the filtered parallel data and evaluate it on the WMT2019 low-resource parallel corpus filtering shared task.

\begin{figure}[t]
    \centering
      \includegraphics[width=0.48\textwidth]{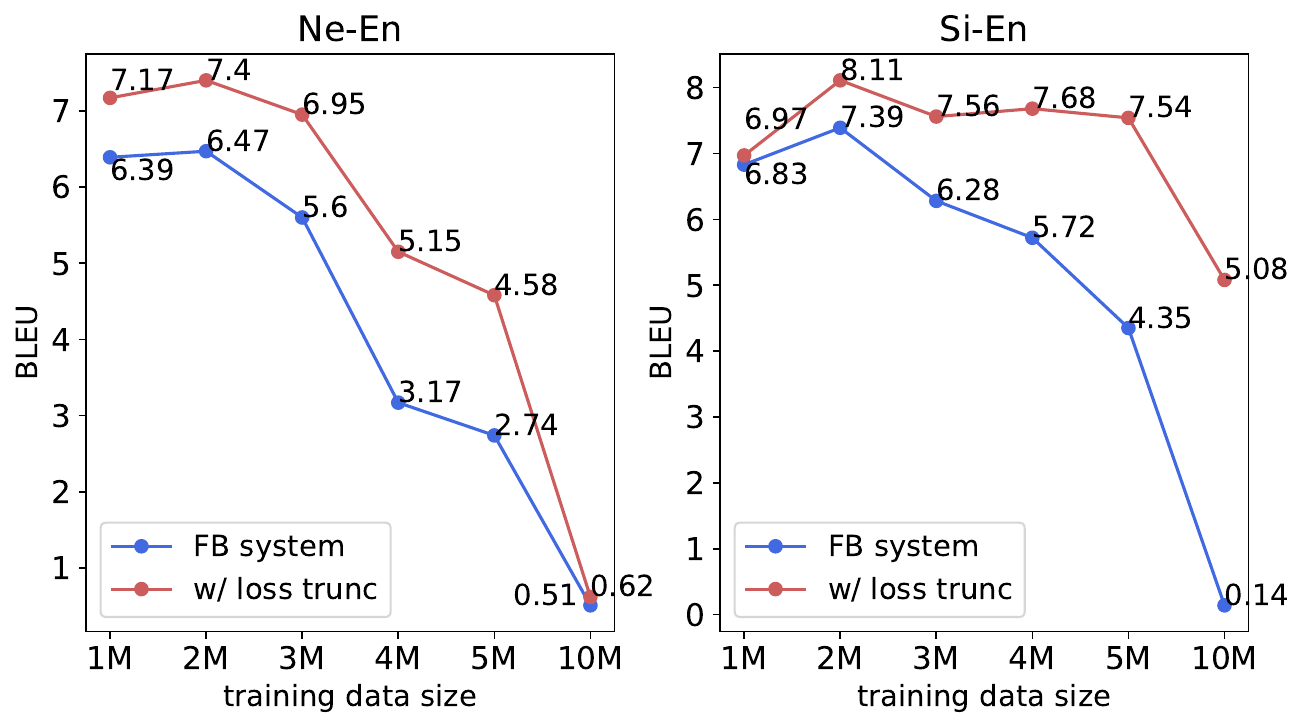}
    \caption{The BLEU scores of the best submission (FB system) in the WMT19 shared task on parallel noisy corpus filtering and our method (w/ loss trunc) on the Ne-En and Si-En \textsf{flores} test sets.}
    \label{fig:exp:wmt19}
    \vspace{-4mm}
\end{figure}
\paragraph{Experimental Setup} The WMT19 shared task focuses on two low-resource languages -- Nepali and Sinhala. It released a very noisy 40.6 million-word (English token count) Nepali-English and a 59.6 million-word Sinhala-English corpus crawled from the web. Participants were asked to score each sentence pair in the noisy parallel set. Scores were used to subsample sentence pairs amounting to 1 million and 5 million English words, which were used to train an MT system that was evaluated on the test set using SacreBLEU~\citep{post-2018-call}. In addition, the shared task also provides additional clean parallel data for Nepali-English (564K), Sinhala-English (646K) and Hindi-English (1.6M), but they can not be used for training the final NMT system.

First, we train a token-level hallucination prediction system with the combined parallel data from all the three language pairs (as Hindi is related to Nepali). 
Second, we use the scores~\citep{chaudhary2019low} that achieve the best overall performance for both language pairs among all the submissions to select the top-scored 1M, 2M, 3M, 4M, 5M, and 10M data (in English tokens) and predict the token-level hallucination labels on the target side.
We follow the same setup and use the script provided by the shared task to train the NMT model with the selected subsets. During training, we discard losses of tokens that are predicted as hallucinations and only compute the losses for the remaining tokens.
We use the validation and test data from the \textsf{flores} dataset~\citep{guzman2019flores} during training and evaluation.

\paragraph{Results and Analysis}
In Fig.~\ref{fig:exp:wmt19}, we present the BLEU of the best submission (FB system) and our method on the Ne-En and Si-En test sets of the \textsf{flores} dataset. First, with token-level loss truncation, our model achieves the new best results on the \textsf{flores} test set in this shared task for both Ne-En (7.4) and Si-En (8.11). Second, for both language pairs our method further improves the state-of-the-art system when varying the training data sizes. Notably, in the extreme case of 10M training data, which is very noisy, the baseline can not obtain decent BLEU scores for Si-En while our method still achieves reasonable performance (0.14 vs. 5.18). However, for Ne-En data sizes after 2M causes performance of both the baseline and our method to drop significantly, possibly because the dataset contains many pairs of misaligned sentences (the source is not Nepali and the target is not English).
\section{Conclusions}
This work proposed a new task of token-level hallucination detection, created human-annotated benchmark datasets, proposed a method for unsupervised learning of hallucination detectors, and showed that the models can be used to define fine grained losses that improve MT training. 
We demonstrate the remark performance of the proposed hallucination detection method in several downstream tasks, including word-level quality estimation and noisy neural machine translation.
In the future, we hope to create a large-scale pretrained hallucination detector for any dataset or model, and also would extend our method to data-to-text generation scenarios. 
We are also interested in investigating how to leverage our detection methods to mitigate hallucination problems in conditional sequence generation.

\section*{Acknowledgements}
The work in this paper was supported in part by a Facebook SRA Award.
We thank the anonymous reviewers for the insightful feedback that helps us improve the paper.

\bibliographystyle{acl_natbib}
\bibliography{acl2021}

\clearpage
\appendix
\section{Human Evaluations}
\label{appex:eval}
\paragraph{Setup} 
We asked three bilingual speakers to annotate the Chinese-to-English evaluation set $\mathcal{D}_{eval}$, which is composed of 150 sentences from the test set of Zh-En multi-domain dataset~\citep{wang2020aaai}, and 100 sentences from the COVID-19 translation benchmark dataset~\citep{anastasopoulos2020tico} -- TICO.
TICO contains 6 finegrained domains including \textit{Wikisource, Wikivoyage, Wikinews, CMU, PubMed} and \textit{Wikipedia}. we randomly sample 25 examples from each of the four domains -- Wikisource, Wikinews, CMU and PubMed, use these 100 samples for evaluation. We train two varieties of models: a standard base Transformer Seq2Seq model and a model that finetunes the MBART~\citep{liu2020multilingual} model on the training data from $\mathcal{D}_{train}$.
In the human evaluation, three bilingual annotators were presented the Chinese source sentence, the English reference sentence and the MT model generation.
\paragraph{Annotation Guidelines and Process}
We conducted the pilot study and practice sessions with annotators before annotating the final blind test set $\mathcal{D}_{eval}$.
The pilot study was performed on a different evaluation set and we performed analysis on them. Then we conducted an education session with evaluators to make sure that they can fully understand and follow the guidelines.
We find that it is important to define a clear workflow for annotators to execute. 
In the final evaluation, we ask each annotator to read the tokens in the sentence carefully and check if they can be supported by the source sentence in the following order:

(1) If there are tokens (or the entire sentence) that cannot be supported by the source, label all the span(s) with color and mark the sentence as a hallucinated one;

(2) If the annotator can not understand the entire translation, mark the sentence as incomprehensible;

(3) If all the tokens in the translation can be entailed from the source, mark the sentence as a faithful one.

We shuffled the order of sentences so that annotators did not know which translation model was used (TranS2S or MBART).
Besides, we made out the following guidelines to help annotators identify hallucinated spans and distinguish bad translations from hallucinated ones:
(1) If a machine generation contains hallucinations, we ask annotators to minimally mask spans of words as hallucinations such that deleting these spans or replacing these spans with other words can dehallucinate the generation (make the generation a faithful one to the source input). For example, if $T=$``John likes Mary, but Mary does not like John.'' and $G=$``John likes Mary, and Mary likes John.'', ``and" and ``likes" in the latter part of $G$ should be marked as hallucinations. 
(2) We ask annotators not to consider the domain of sentences when marking hallucinations. For examples, if $S$=``\begin{CJK*}{UTF8}{gbsn}今天我的胸部非常痛。
    \end{CJK*}" (Chinese), $T$=``My chest hurts badly today." and $G$=``My breast hurt badly today.", in this case, both the reference $T$ and the MT $G$ are valid translations of the source sentence because the word ``\begin{CJK*}{UTF8}{gbsn}胸部
    \end{CJK*}" in the source is a polysemy. Without considering the domain that sentences come from, the generation is a faithful one.
(3) We ask annotators not to be ``harsh", e.g. if a capitalized word in the reference is lowercased in the translation, we ask them not to mark it as hallucination under the rule that hallucinations should only be considered by the meaning of words and whether they are faithful to the source, instead of the surface form.

Note that annotations are performed on the raw sentences, i.e. punctuation marks can also be labeled as hallucinations along with the span and we did not apply special treatments to them. At test time, the model outputs are compared against the raw form of sentences, and model predictions on subwords are converted to labels on the raw sentences.
Besides, based on our guidelines, the annotated span of hallucination words may also contain prepositions and other stop words.
\vspace{-2mm}
\paragraph{Post-processing: }
We dropped all the translations that were labeled as incomprehensible (15 for TranS2S and 3 for MBART).
To aggregate annotations from the three annotators, we assign the label to each token by majority voting, i.e. the label that two or more annotators agree on. We also aggregate the evaluation data from~\citet{maynez_acl20} in the same manner to produce our own test set for abstract text summarization.

\begin{table}[h]
     \centering
     \small
    \begin{tabular}{lcc}
        \toprule
   \multirow{2}{*}{\textbf{Models}}     & \multicolumn{2}{c}{\textbf{Fleiss' Kappa}} \\
         & \textbf{Token} &\textbf{Sent} \\
        \midrule
        MT \\
        \;\; TranS2S & 0.58 & 0.72 \\
        \;\; MBART  & 0.54 & 0.62  \\
        \midrule
        XSum \\
        \;\; PtGen & 0.81 & -  \\
        \;\; TConvS2S & 0.83 & - \\
        \;\; TranS2S & 0.79 & -  \\
        \;\; BERTS2S & 0.79  & - \\
        \bottomrule
    \end{tabular}
    \caption{\label{tab:fleiss} Fleiss's Kappa scores ($\uparrow$): agreements on \textbf{token}-level hallucination labels or sentence-level (\textbf{sent}) ratings among different annotators. The token-level agreements for \textsc{XSum} are computed on the released annotations by \citet{maynez_acl20}.}
    \vspace{-6mm}
\end{table}


 \section{Training of NMT models}
\label{appex:exp}
\paragraph{Tokenization} For TranS2S, we first segment the Chinese corpus with a Chinese word segmentation tool~\citep{pkuseg}, then we learn separate BPE vocabularies with 32k merge operations~\citep{sennrich2015neural} over the source (Zh) and the tokenized target (En) corpus respectively. For MBART, we directly apply the contained sentence-piece dictionary in the finetuned model to the raw data of Chinese and English corpus.
\paragraph{Model} We use the implementation of Transformer from fairseq~\citep{ott2019fairseq}. Following the notations used in fairseq, we use a base transformer model for TranS2S and a large tranasformer model for MBART.
\paragraph{Training and Decoding} For TranS2S, we apply the standard hyperparameters reported in the example of fairseq. We use the Adam optimizer~\citep{kingma2014adam} using $\beta_1=0.9, \beta_2=0.98, \epsilon=1e-8$. The learning rate is scheduled using \texttt{inverse\_sqrt} with a maximum learning rate $0.0005$ and $4000$ warmup steps. We set the label smoothing as $0.1$. We apply dropout of 0.1 and select the best model with validation BLEU scores.
We run the model on $8$ GPUs for $300,000$ updates with an effective batch size of around $64,000$ tokens.
When finetuning MBART, we use learning rate of 3e-5, and use \texttt{polynomial\_decay} for learning rate scheduling with warmup updates of 3,000. The effective batch size is 16,384. Dropout is set to be 0.3 and the attention dropout rate is 0.1. The label smoothing is set to be 0.2. We finetune MBart for 60,000 updates.
We decode outputs with beam-search and beam size of 5.

\section{Experimental Details for Token-level Hallucination Prediction}
\label{app:exp:finetune}
\paragraph{Subword Tokenization}
Depending on the pretrained model (Roberta / XLM-Roberta) we finetune on, we apply corresponding subword segmentation to the synthetic data set ($S, T, T'$) and calculate the edit-distance between the $T$ and $T'$ at the subword level.
At evaluation time, the model predicts the hallucination labels for each subword in the sentence, thus we predict a word to be a hallucination word if any subword of it is predicted as a hallucinated one.

\paragraph{Synthetic data generation} There are a couple of hyperparameters of noised functions in the BART implementation~\citep{ott2019fairseq}. The main noised functions include (1) random masking, (2) random replacement, (3) random insertion of masks. We found that random masking and random replacement are the two key factors affecting the generated sentences and we have provided their settings in the main paper. We apply a random insertion masks rate of 0.2 for all settings. In addition, the noise functions are applied to words instead of spans in our setting. 

\paragraph{Finetuning} For MT, we finetune a large XLM-Roberta~\citep{conneau-etal-2020-unsupervised} released in fairseq~\citep{ott2019fairseq}.
For summarization, we finetune a large Roberta~\citep{ott2019fairseq} on the synthetic data where we truncate articles that exceed 512 tokens (allowed by the Roberta) to be 512. 
For both models, we use the Adam optimizer~\citep{kingma2014adam} with $\beta_1=0.9, \beta_2=0.98, \epsilon=1e-6$ and weight decay of 0.1.
We set the masking probability to be 0.35 for the $\mathcal{L}_{mlm}$ loss. The dropout and attention dropout rates are set to be 0.1.
We adopt \texttt{polynomial\_decay} for learning rate scheduling with learning rate of 2e-5.

\begin{table}[t]
\small
\centering
\setlength{\tabcolsep}{2pt}
    \begin{tabular}{lccc}
        \toprule
        \textbf{Input to $\mathcal{N}(\cdot)$} & \textbf{Precision} &\textbf{Recall} & \textbf{F1}\\
        \midrule
        MT \\
        \;\; raw & 58.35 & 70.12 & 63.70 \\
        \;\; TranS2S distill  & 64.27 & 67.30	& 65.75 \\
        \midrule
        Summarization \\
        \;\; raw & 57.02 &	67.23 &	61.70 \\
        \;\; Extractive distill & 54.10	& 36.45 & 43.55 \\
        \;\; Abstractive distill & 57.33 & 28.59 & 38.16 \\
        \bottomrule
    \end{tabular}
    \caption{\label{tab:distill} Performance on the TranS2S benchmark from MT and summarization by using different data as the input to the noised function $\mathcal{N}(\cdot)$. ``raw" refers to the original targets in the training data.}
    \vspace{-4mm}
\end{table}
\begin{figure}[h]
  \centering
    \includegraphics[width=0.4\textwidth]{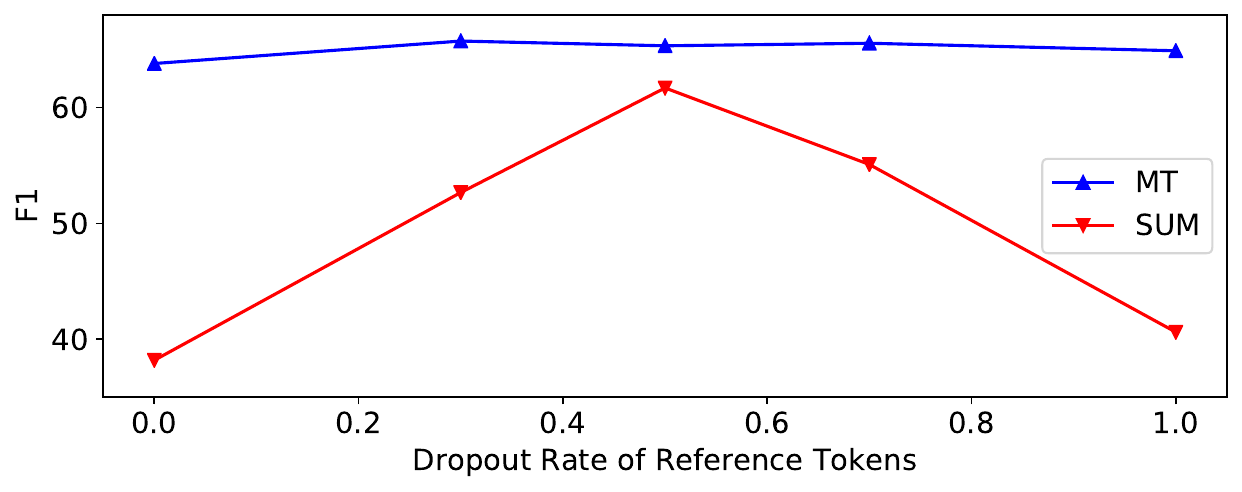}
  \caption{\label{fig:dropref}Performance on the TranS2S outputs from MT and summarization by varying the token dropout rate of in the reference at training time.}
  \vspace{-5mm}
\end{figure}
\begin{table}[t]
\centering
 \setlength{\tabcolsep}{2pt}
\small
\begin{tabular}{lcc}
\toprule  
  \textbf{Methods}
 & \textbf{TranS2S} & \textbf{MBART}\\
\midrule
Alignment & (18.90, 66.82, 29.47) & (5.63,	42.09,	9.93) \\
Overlap-based & (7.02,	13.10,	9.14) & (1.98,	8.97,	3.24)  \\
Synonym-based & -- & --  \\
\midrule
Ours (w/o ref) & \textbf{(64.27, 67.30, 65.75)} & \textbf{(49.56, 36.32, 41.92)} \\
Ours (w/ ref) & (59.92, 74.27, 66.08) & (43.13, 53.63, 46.81)\\
\bottomrule
\end{tabular}
\caption{Triplets represent (Precision, Recall, F1 (x100)) of hallucination labels on the outputs of different systems from a MT task (\S\ref{sec:data:mt}). The first block are baseline methods and the second block are our results. We highlight the best results without using reference.}
\label{tab:token:full:mt}
\end{table}

\begin{table}[h]
\vspace{-1mm}
\centering
\small
\begin{tabular}{p{2cm}p{5cm}}
\toprule
\textbf{Source} & 
\begin{CJK*}{UTF8}{gbsn}
信息组 被 称作 页面 数据。
\end{CJK*}\\
\textbf{Reference} & the set of information is called page data. \\
\textcolor{orange}{\textbf{Generation}} & the foreign[1] mix[1] is called the page data.\\
\midrule
\textbf{Source} & 
\begin{CJK*}{UTF8}{gbsn}
金属线 对应 于 第一 电阻器。
\end{CJK*}\\
\textbf{Reference} & the metal lines correspond to first resistors.\\
\textcolor{orange}{\textbf{Generation}} & the wire corresponds with the first capital[1]. \\
\midrule
\textbf{Source} & 
\begin{CJK*}{UTF8}{gbsn}
驱动 样本 流过 液流 通路;
\end{CJK*}\\
\textbf{Reference} & driving samples to flow through a flow channel; \\
\textcolor{orange}{\textbf{Generation}} & driving samples pass the flow of people[1];\\
\bottomrule
\end{tabular}
\caption{Examples of partially hallucinated outputs from the teacher MT model used in self-training and the hallucinated labels predicted by our system. We only highlight words with hallucination labels with [1].}
\label{tab:st:example}
\vspace{-4mm}
\end{table}

\begin{table*}[h]
\centering
\small
\begin{tabular}{lcccc}
\toprule  
  \textbf{Methods}  & \textbf{PtGen} & \textbf{TConvS2S} & \textbf{TranS2S} & \textbf{BERTS2S} \\
\midrule
Alignment &  (60.66,	28.65,	38.92) & (66.14,	26.60,	37.94) & (56.24,	24.85,	34.47) & (50.68,	27.69,	35.81) \\
Overlap-based & (67.72,	49.54,	57.22) & (60.39,	49.24,	54.25) & (53.22,	54.37,	53.79) & (62.57,	49.26,	55.13)  \\
Synonym-based & (50.52,	72.50,	59.55) & (57.06,	72.16,	63.73) & (50.29,	70.37,	58.66) & (41.80,	72.67,	53.07) \\
\midrule
Ours (w/o ref) & (57.47, 71.35, 63.66) & (63.21, 68.93, 65.94)	& (57.02, 67.23, 61.70) & (49.83, 62.50, 55.45) \\
Ours (w/o ref + syn) & \textbf{(50.33, 90.27, 64.72)} & \textbf{(56.86, 88.93, 69.37} & \textbf{(50.21, 87.78, 63.88)} & \textbf{(41.70, 87.52, 56.49}) \\
Ours (w/ ref) & (56.51, 73.48, 63.89) & (61.68, 71.63, 66.28) &	(55.88, 70.19, 62.24) &	(48.39, 66.11, 55.88)\\
\bottomrule
\end{tabular}
\caption{Triplets represent (Precision, Recall, F1 (x100)) of hallucination labels on the abstract summarization task (\textsc{XSum} dataset). The first block are baseline methods and the second block are our results. We highlight the best results without using reference.}
\vspace{-2mm}
\label{tab:token:full:xsum}
\end{table*}

\begin{table*}[h]
\centering
\small
\begin{tabular}{p{2.5cm}p{10cm}}
\toprule
\textcolor{blue}{Reference} & \textcolor{blue}{the arrangement pattern of the projections 2 will now be explained with reference to figs. 5-7.} \\
Annotation & 
next,[0] we[0] use[0] fig.[0] 5[0] -[0] 7[0] to[0] explain[0] the[0] disposition[0] pattern[0] with[0] pm-2.[1] \\
Prediction &  
next,[0] we[0] use[0] fig.[0] 5[0] -[0] 7[0] to[0] explain[0] the[0] disposition[0] pattern[0] with[1] pm-2.[1] \\
\midrule
\textcolor{blue}{Reference} & \textcolor{blue}{a swivel joint 557 is provided in a radially outer region, on an end surface of the drive plate 556.}\\
Annotation & a[0] rotation[0] hinged[1] 557[0] is[0] provided[0] to[0] the[0] external[0] area[0] on[0] a[0] trail[1] that[0] has[0] a[0] preface[1] state.[1] \\
Prediction & 
a[0] rotation[0] hinged[0] 557[0] is[0] provided[1] to[0] the[0] external[0] area[0] on[0] a[0] trail[1] that[1] has[1] a[0] preface[1] state.[1]\\
\midrule
\textcolor{blue}{Reference} & \textcolor{blue}{if you have a fever of a hundred and two or higher.}\\
Annotation & if[0] your[0] heat[0] reaches[0] 102.d[0] egree.[0] f.[0] or[0] above,[0] \\
Prediction & if[0] your[0] heat[0] reaches[0] 102.d[1] egree.[1] f.[1] or[0] above,[0]\\
\bottomrule
\end{tabular}
\caption{Examples of annotations and our hallucination detection model predictions, [0] and [1] respectively indicate faithful and hallucinated word.}
\label{tab:pred:examples}
\vspace{-4mm}
\end{table*}

\begin{figure*}[h]
\vspace{-3mm}
    \centering
      \includegraphics[width=1.0\textwidth]{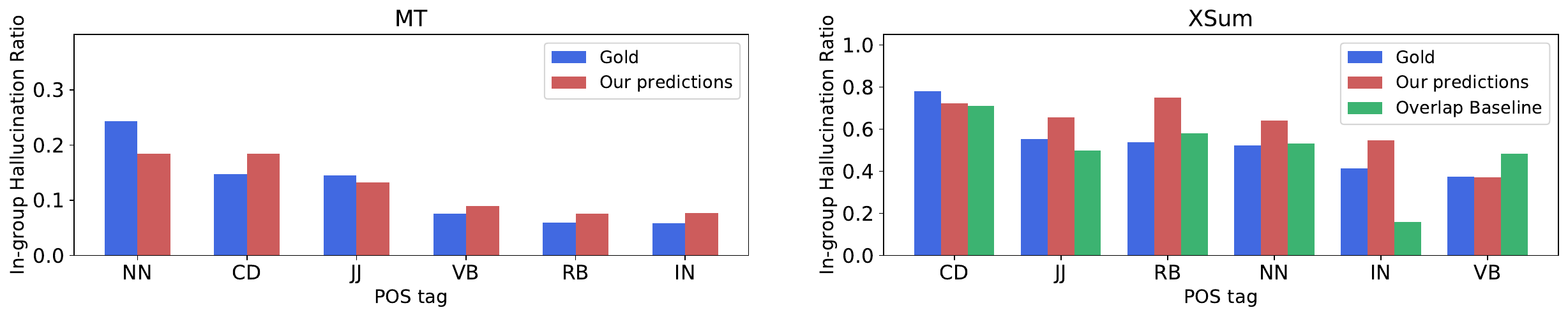}
    \caption{Analysis of part-of-speech tags and within-group percentage of hallucinations for MT (left) and summarization (right).}
    \label{fig:micro:pos}
\vspace{-3mm}
\end{figure*}
\section{Ablation Studies}
\label{sec:abl}
\paragraph{Effects of including reference at training time}
Recall that we concatenate the source, reference and machine generation together as the input when learning hallucination predictions (Sec.~\ref{sec:model:finetune}).
In Fig.\ref{fig:dropref}, we vary the dropout rate of tokens in the reference at training time and evaluate the models on the outputs from the TranS2S model for both tasks, where dropout rate of 1.0 indicates that we do not include the reference at all. \textbf{First}, different dropout rates do not signficinatly affect performance for MT, this is likely because we use the paraphrased target when creating the synthetic data instead of the reference sentences. Thus, the ``hallucinated" sentences $D'$ from BART do not resemble the reference $T$ as closely as $T'$, and the model will not learn spurious correlations between the $T$ and $D'$. 
\textbf{Second}, for summarization we see that applying word dropout is crucial since we have used the reference more directly for generating synthetic data. On the other hand, if reference is removed at learning time (dropout = 1.0), the resulted model performs poorly, which shows that including reference at training time also has positive effects.

\vspace{-2mm}
\paragraph{Effects of paraphrased data}
We investigate the effects of using paraphrased data in Tab.~\ref{tab:distill}, where we apply the noise functions to different forms of targets when generating synthetic data. For MT, we create paraphrased targets via knowledge distillation~\citep{kim2016sequence} where we use the output from TranS2S conditioned on the source sentence in the bi-text corpus as the paraphrased target.
We can see that with distillation data for synthetic data generation, the model achieves better results compared to using the references. However, note that we need to choose a proper word dropout rate when using the reference-based synthetic data as discussed above.
For abstractive summarization, we create paraphrased data out of an abstractive and an extractive summarization systems respectively. We finetune BART on the bi-text of \textsc{XSum} and create distillation data from this finetuned abstractive model. For the extractive system, we use the recent proposed MatchSum~\citep{zhong2020extractive} as the distillation model. We see a significant drop in the performance for both of the variants. This likely due to the fact that: (1) it has been shown that abstractive summarization systems are prone to hallucinate contents themselves~\citep{maynez_acl20}, thus we are not able to create reliable pseudo labels based on the generated summaries, and (2) the extractive system generates summaries out of the input article which diverge from the actual abstractive summaries we evaluate on, and the model cannot generalize well under such data shift.

\vspace{-2mm}
\section{Supplymental Results and Analysis}
\vspace{-1mm}

\vspace{-1mm}
\subsection{Full Results of Token-level Hallucination Predictions}
\label{app:full:results}
We found the synonym and string-matching based methods are strong and effective baselines on the monolingual (summarization) token-level hallucination prediction task as an alternative to neural methods.
However, previous work~\citep{maynez_acl20,wang-etal-2020-asking,durmus-etal-2020-feqa} on hallucination assess did not study synonym-based non-neural baselines when measuring the faithfulness of the summarization model outputs. 

\vspace{-1mm}
\subsection{Analysis on Part-of-speech tags and with-in Group Hallucination Percentage}
\label{app:micro:pos}
We have shown that the macro Part-of-Speech tag distribution of hallucinated tokens in \S\ref{sec:analysis}. In this section, we analyze the micro-percentage of hallucination labels within each POS tags. 
We show the gold annotations as well as our model predictions of hallucination words within each POS tags. For summarization, we also show the results from the string-matching baseline.
From Fig.~\ref{fig:micro:pos}, we can see that for MT nouns are most likely hallucinated words while for summarization cardinal numbers (e.g. \textit{one, two}) are most likely hallucinated words.
And we can see that our model predictions align well with the gold annotations on the percentage of hallucinated words within each POS tags.

\vspace{-1mm}
\subsection{Examples of Partially Hallucinated Outputs from Teacher MT Model}
\label{app:st:examples}
In Tab.~\ref{tab:st:example}, we randomly select some examples for which we present the source sentences from the patent monolingual Chinese dataset, the corresponding reference English sentences and the generations from a teacher model trained on the training data described in \S\ref{sec:data:mt} where patent is a low-resource domain. We can see that in these examples, only parts of the model outputs are hallucinated and the rest of the outputs are good translations that are faithful to the source. Through our approach in \S\ref{sec:exp:st}, we can still make use of these good parts of translation during training.

\vspace{-1mm}
\subsection{Examples of Hallucination Predictions on the MT test set}
\label{app:pred:examples}
As shown Tab.~\ref{tab:pred:examples}, our model performs well in general but can be inaccurate in case of spelling errors of the translations. Besides, we also find some annotation errors while our model predicts correctly.

\end{document}